\documentclass[letterpaper]{article} 
\usepackage{aaai2026}  
\usepackage{times}  
\usepackage{helvet}  
\usepackage{courier}  
\usepackage[hyphens]{url}  
\usepackage{graphicx} 
\usepackage{natbib}  
\usepackage{caption} 
\usepackage{algorithm}
\usepackage{algorithmic}
\usepackage{amsmath}
\usepackage{xspace}
\usepackage{multirow}
\usepackage{subfig}
\usepackage{booktabs}
\usepackage{amssymb}
\usepackage{algorithmic}
\usepackage{bbding}
\usepackage[dvipsnames]{xcolor}

\usepackage{newfloat}
\usepackage{listings}
\DeclareCaptionStyle{ruled}{labelfont=normalfont,labelsep=colon,strut=off} 
\floatstyle{ruled}
\newfloat{listing}{tb}{lst}{}
\floatname{listing}{Listing}
\title{ThinkFake: Reasoning in Multimodal Large Language Models for AI-Generated Image Detection}

\author{
    Tai-Ming Huang$^{1,2}$, Wei-Tung Lin$^{2,5}$, Kai-Lung Hua$^{4,5}$, \\ Wen-Huang Cheng$^{1}$, Junichi Yamagishi$^{3}$, Jun-Cheng Chen$^{2}$
}
\affiliations{
    National Taiwan University\textsuperscript{\rm 1}, Academia Sinica\textsuperscript{\rm 2}, National Institute of Informatics\textsuperscript{\rm 3}, \\ Microsoft\textsuperscript{\rm 4} National Taiwan University of Science and Technology\textsuperscript{\rm 5}
    
}

\usepackage{bibentry}

\begin{document}

\maketitle

\begin{abstract}
The increasing realism of AI-generated images has raised serious concerns about misinformation and privacy violations, highlighting the urgent need for accurate and interpretable detection methods. While existing approaches have made progress, most rely on binary classification without explanations or depend heavily on supervised fine-tuning, resulting in limited generalization. In this paper, we propose ThinkFake, a novel reasoning-based and generalizable framework for AI-generated image detection. Our method leverages a Multimodal Large Language Model (MLLM) equipped with a forgery reasoning prompt and is trained using Group Relative Policy Optimization (GRPO) reinforcement learning with carefully designed reward functions. This design enables the model to perform step-by-step reasoning and produce interpretable, structured outputs. We further introduce a structured detection pipeline to enhance reasoning quality and adaptability. Extensive experiments show that ThinkFake outperforms state-of-the-art methods on the GenImage benchmark and demonstrates strong zero-shot generalization on the challenging LOKI benchmark. These results validate our framework’s effectiveness and robustness. Code will be released upon acceptance.
\end{abstract}

\section{Introduction}
\label{sec:intro}
The rapid development of generative models~\cite{dalle, dhariwal2021adm, li2024stylegan} has led to AI-generated images that are increasingly realistic and difficult to distinguish from real ones. While these advancements benefit creative industries, they also raise serious concerns regarding misuse, including misinformation, fraud, and privacy violations. As techniques such as GANs~\cite{gan}, diffusion models~\cite{esser2024sd3}, and flow-based approaches~\cite{rombach2022ldm, flux2024} continue to evolve, the challenge of detecting synthetic content grows more severe. This trend highlights the urgent need and importance of detection methods that are both accurate and interpretable.

Recent studies~\cite{wang2019cnnspot, yan2025aide, npr, ojha2023univfd, antifakeprompt} have shown significant progress in detecting AI-generated images. However, most existing detection models are limited to binary classification and lack meaningful explanations, which makes their outputs difficult to interpret and unreliable in real-world scenarios. To improve interpretability, some recent approaches~\cite{fakereasoning, zhou2025aigi, ji2025towards, kang2025legion,wen2025fakevlm} leverage Multimodal Large Language Models (MLLMs) due to their strong capabilities in commonsense reasoning and natural language generation. Despite their promise, these methods often rely on large-scale annotated datasets for supervised fine-tuning. This heavy dependence on labeled data leads to memorization rather than genuine reasoning, ultimately resulting in poor generalization. While these advancements are promising, two key challenges remain: \textbf{1) lack of interpretability:} most models provide only binary outputs without explaining their decisions, which limits their practical applicability; and \textbf{ 2) lack of reasoning and generalization:} methods that incorporate MLLMs for explanation typically rely on extensive supervision and lack autonomous reasoning, making it difficult to generalize to complex, unseen scenarios. Therefore, developing AI-generated image detectors that are both capable of reasoning and generalization has become increasingly urgent. ``Reasoning'' in MLLMs represents a major step forward in handling complex analytical tasks. Unlike traditional classifiers that rely on pattern matching or likelihood scores, reasoning-capable MLLMs can produce step-by-step explanations that resemble human cognition. The principle of ``SFT memorizes, RL generalizes'' has been applied in recent work~\cite{chu2025sft}, raising an important question: can we leverage this reasoning ability not only to detect AI-generated images, but also to explain decisions through logic, comparison, and context?

To address the above challenges, we propose a reasoning-based and generalizable framework for AI-generated image detection, inspired by recent progress in reasoning tasks~\cite{segzero, thinkact} and reinforcement learning~\cite{GRPO}. Our method builds on a reasoning MLLM, guided by carefully designed prompts and reward functions, to enhance both performance and generalization.
We further introduce a structured detection pipeline combined with Group Relative Policy Optimization (GRPO) reinforcement learning to strengthen the model’s reasoning and adaptability. Through extensive experiments and ablation studies, our approach outperforms existing state-of-the-art methods on the GenImage~\cite{zhu2024genimage} benchmark. Moreover, on the more challenging and comprehensive LOKI~\cite{ye2024loki} benchmark, our method demonstrates strong zero-shot generalization, validating the effectiveness of our framework.
Our main contributions are summarized as follows:
\begin{itemize}
\item We develop a novel reasoning MLLM framework that not only achieves strong detection performance but also provides interpretable reasoning for AI-generated image detection.
\item We design specialized prompts and reward functions that significantly enhance the model's detection capability, performance, and explainability.
\item Our method outperforms state-of-the-art approaches on the GenImage benchmark. Extensive experiments further demonstrate the framework's generalization ability and robustness under real-world challenges.
\end{itemize}

\section{Related Work}
\label{sec:background}

\begin{figure*}[t]
    \centering
    \includegraphics[width=\textwidth]{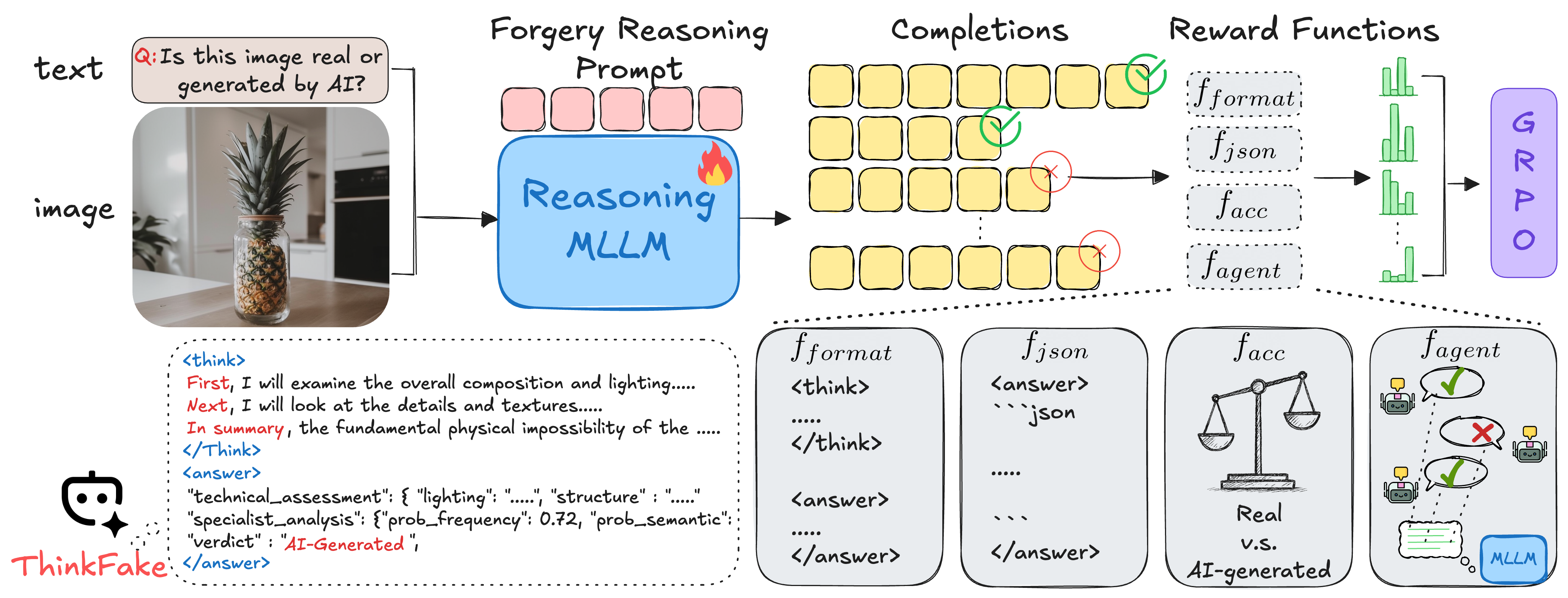}
    \caption{\textbf{Overview of our ThinkFake:} We propose a reasoning MLLM framework capable of deliberate reasoning for detecting AI-generated images. Guided by a forgery reasoning prompt, and trained with carefully designed reward functions and GRPO-based reinforcement learning, the model demonstrates strong generalization and self-reflective explanatory capabilities.}
    \label{fig:system} 

\end{figure*}
\begin{figure*}[t]
    \centering
    \includegraphics[width=\textwidth]{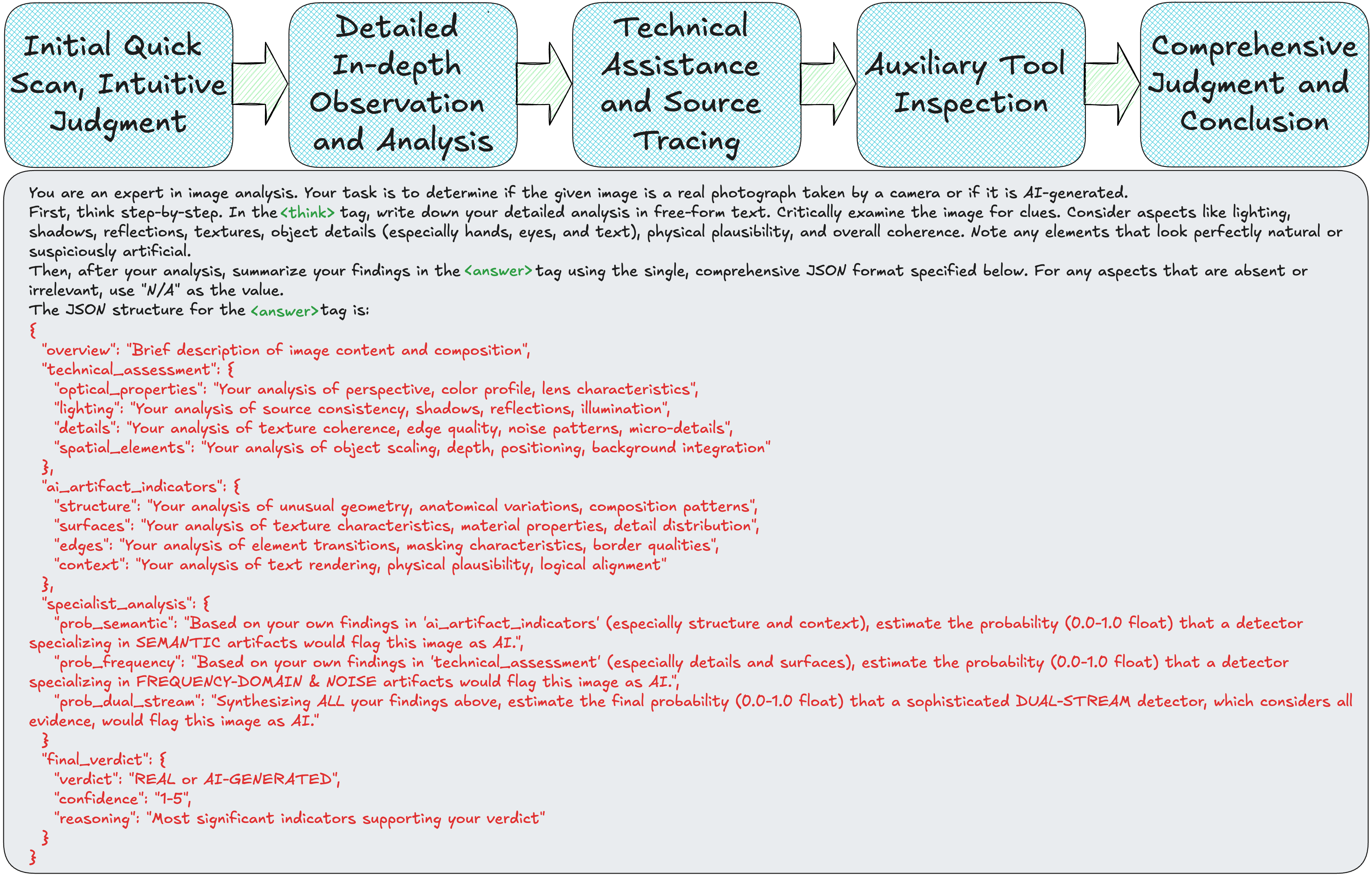}
    \caption{Top: Overview of ThinkFake detection pipeline. Bottom: Detail of our proposed \textit{Forgery Reasoning Prompt}.}
    \label{fig:pipeline} 
    \vspace{-10pt}
\end{figure*}

\subsection{AI-Generated Image Detection}
The rapid advancement of generative AI models has intensified the challenge of detecting AI-generated images, as recent models including MidJourney~\cite{midjourney}, DALLE-3~\cite{dalle}, Stable Diffusion~\cite{stablediffusion}, and Flux~\cite{flux2024} produce increasingly realistic images. FreDect~\cite{frank2020fredect} identifies anomalies by analyzing frequency domain characteristics of GAN-generated images. UnivFD~\cite{ojha2023univfd} leverages pre-trained CLIP-ViT features with nearest neighbor and linear probing for cross-domain generalization. DIRE~\cite{wang2023dire} exploits reconstruction errors to distinguish diffusion-generated images from real ones, though it suffers from poor generalization and high computational overhead. PatchCraft~\cite{zhong2023patchcraft} segments images into patches and applies SRM filters to analyze pixel correlations for detection. AEROBLADE~\cite{ricker2024aeroblade} introduces a training-free approach using autoencoder reconstruction errors for latent diffusion model detection. AIDE~\cite{yan2025aide} employs a dual-stream architecture combining frequency and semantic features. However, these approaches often fail to generalize to out-of-distribution data due to limited training data and the lack of prior knowledge from Multimodal Large Language Models (MLLMs)~\cite{bai2025qwen2,liu2023visual}. In addition, their binary predictions lack explanations, limiting their practical applicability.


\subsection{Explainability}


Explainability in MLLMs is crucial for mitigating risks associated with AI misuse, and recent research has increasingly investigated this domain. Some studies~\cite{sida, canchatgptdetect} directly prompt MLLMs with queries such as ``explain what the artifacts are.'' However, research reveals that directly generating textual explanations frequently results in hallucinations or overthinking, leading to inaccurate outcomes or refusal to respond~\cite{canchatgptdetect}.
Furthermore, MLLMs often fail to comprehensively perceive all relevant features, which constrains their explainability effectiveness. To overcome these limitations, researchers have employed fine-tuning approaches such as LoRA~\cite{lora} or Direct Preference Optimization (DPO)~\cite{dpo} for MLLMs~\cite{huang2024ffaa,wen2025fakevlm,fakereasoning,zhou2025aigi}. However, these methods tend to memorize training patterns rather than generalize to real-world scenarios. To address this limitation, we employ GRPO~\cite{deepseekr1}, which enhances generalization by letting the model learns how to ``think'', for improved MLLM detection and explainability.

\section{Method}
\label{sec:method}


\subsection{Overview}\label{sec:overview} 
In this paper, we aim to train a model that effectively distinguishes AI-generated images from real images captured by cameras. Leveraging recent advancements in multimodal large language models (MLLMs)~\cite{chen2024internvl, bai2025qwen2}, we utilize their capabilities in image understanding and text generation to analyze image authenticity and provide reliable explanations. We propose ThinkFake, a carefully designed framework that employs a reasoning MLLM $\mathcal{F}_{\theta}$ with a reinforcement learning process to classify and reason about whether an image is real or AI-generated. 
Given an input image $\mathrm{I}$ and a corresponding prompt $\mathrm{P}$, with a simple question \textit{ ``is this image real or generated by AI?''}, the reasoning MLLM $\mathcal{F}_{\theta}$ produces both verdict $\mathrm{V}$ and reasoning explanation $\mathrm{R}$. This process can be represented by the following equations:
\begin{align}
\label{eq:eq_overview_2}
\mathrm{V}, \mathcal{R} &= f_{post}({\mathcal{F}_{\theta}}(\mathrm{I}, \mathrm{P})),
\end{align}
where $f_{\text{post}}(\cdot)$ denotes the post-processing function. 

Specifically, we developed a comprehensive detection pipeline and training process for the MLLM. First, we designed a task-specific \textit{Forgery Reasoning Prompt} tailored for the MLLM to address this task. Inspired by~\cite{deepseekr1}, we then employed reinforcement learning with four carefully designed reward functions to enhance the MLLM’s reasoning capabilities for this task, enabling it to perform both detection and explanation, as illustrated in Fig.~\ref{fig:system}.

\subsection{Pipeline and Forgery Reasoning Prompt}\label{sec:pipeline}
\noindent\textbf{Pipeline:} Inspired by~\cite{fakereasoning}, detecting AI-generated content requires a multifaceted approach that examines various visual attributes, such as lighting, shadows, textures, and edges. By analyzing these features, it is possible to identify recurring patterns or artifacts commonly associated with AI-generated content, thereby facilitating robust detection. Drawing on recent advancements in multi-modal chain-of-thought (CoT) reasoning on Multimodal Large Language Models (MLLMs), we propose a novel chain-of-thought pipeline for AI-generated image detection. This pipeline incorporates hierarchical reasoning steps to enhance the visual reasoning capabilities of MLLMs. First, we decompose the detection pipeline into five steps, as shown in the upper part of Fig.~\ref{fig:pipeline}:

\begin{itemize}
\item \textbf{Initial Quick Scan and Intuitive Judgment:} In the first step, an initial impression of the input image is formed, evaluating its overall harmony, content, and composition.
\item \textbf{Detailed In-Depth Observation and Analysis:} Next, a meticulous examination is conducted, focusing on texture quality, lighting, shadows, reflections, edges, and blending, with each aspect analyzed in detail.
\item \textbf{Technical Analysis and Source Tracing:} During the technical analysis and source tracing phase, specific attention is given to artifacts commonly found in AI-generated content, such as inconsistencies in structure, surfaces, edges, and contextual elements.
\item \textbf{Auxiliary Tool Inspection:} Several specialized AI-generated image detection tools, each focusing on different aspects, are available as reference or auxiliary tools to support comprehensive detection and evaluation.
\item \textbf{Comprehensive Judgment and Conclusion:} In the final stage of the pipeline, evidence and analysis results from all previous steps are synthesized to deliver a final judgment and reasoning, completing the comprehensive detection process.
\end{itemize}

\newcommand{\lt}{\textless}
\newcommand{\gt}{\textgreater}

\noindent\textbf{Forgery Reasoning Prompt:} 
Inspired by recent advancements in structured output for MLLMs~\cite{Tam2024let, he2024does}, we unify the analytical process described above and provide clear guidance for evaluating each attribute. As illustrated in the lower part of Fig.~\ref{fig:pipeline}, we introduce the \textit{Forgery Reasoning Prompt} to enhance the reasoning capability of MLLMs, specifically for the task of detecting AI-generated images. Each analytical step is incorporated into the prompt using special tokens, ~\lt think\gt~and~\lt answer\gt, together with a strict JSON format to constrain the model’s output. This structured format improves both interpretability and detection performance. In the ``Auxiliary Tool Inspection'' stage, we define three expert perspectives: a semantic expert that focuses on high-level features, a frequency expert that targets low-level features, and a dual-stream expert that considers both. This design allows the \textit{Forgery Reasoning Prompt} to support comprehensive reasoning and guide the model through a complete detection pipeline, ultimately producing interpretable and structured outputs.

\subsection{Reward Functions}\label{sec:reward}
Reward functions guide model optimization in reinforcement learning. We design four binary rewards, each set to 1 if satisfied and 0 otherwise.

\noindent\textbf{Reasoning Format Reward ($f_{format}$).} The model's output must adhere to the \lt think\gt...\lt /think\gt~and~\lt answer\gt...\lt/answer\gt~format. This reward constrains the model to output its reasoning process within the \lt think\gt~tags and provide the final answer within the \lt answer\gt~tags.

\noindent\textbf{JSON Format Reward ($f_{json}$).} To support Forgery Reasoning Prompt, we impose stricter constraints on the model's response format, requiring answers to adhere to a structured JSON format. This constraint improves detection performance, enhances the efficiency of answer extraction, and provides interpretable outputs.

\noindent\textbf{Accuracy Reward ($f_{acc}$).} To evaluate the model's performance, we implemented a reward function that measures binary accuracy by comparing the predicted answer to the ground-truth label. Specifically, the model receives a reward of 1 when its prediction matches the ground-truth label, and 0 otherwise. This straightforward yet effective scheme offers a clear and reliable signal for guiding the model toward accurate classification.

\begin{table*}[t]
\centering
\resizebox{\textwidth}{!}{
\begin{tabular}{ll|c|*{9}{c}}
\toprule
\textbf{Model} & & \textbf{Params.} &
{Midjourney} & {SD v1.4} & {SD v1.5} &
{ADM} & {GLIDE} & {Wukong} &
{VQDM} & {BigGAN}  & {\textbf{\textit{Mean}}} \\
\midrule
\multicolumn{12}{c}{\textbf{Closed-Source MLLMs}} \\
\midrule
ChatGPT-4o~\cite{openai2024gpt4ocard} & & -- & 56.8 & 75.3 & 75.7 & 65.8 & 79.3 & 84.4 & \textbf{79.7} & \textbf{84.6} & 75.2  \\
ChatGPT-o3~\cite{openai2025gpto3card} & & -- & 62.0 & 81.6 & 81.2 & 68.8 & 80.5 & 91.0 & 79.0 & 81.4 &  78.2 \\
Gemini-2.0-flash~\cite{comanici2025gemini} &  & -- & 55.5 & 80.1 & 77.8 & 65.4 & 66.5 & 88.2 & 78.8 & 79.4   & 73.9 \\
\midrule
\multicolumn{12}{c}{\textbf{Open-Source General MLLMs}} \\
\midrule
Qwen2.5-VL-7B-Instruct~\cite{bai2025qwen2} & & 7B & 54.6 & 51.9 & 50.6 & 52.3 & 58.0 & 56.3 & 54.7 & 54.2 & 54.1 \\
Qwen2.5-VL-72B-Instruct~\cite{bai2025qwen2} & & 72B & 54.5 & 56.0 & 54.9 & 55.5 & 62.3 & 65.1 & 61.1 & 61.6  & 58.9 \\
InternVL2.5-78B~\cite{chen2024internvl} & & 78B & 52.7 & 53.4 & 53.0 & 55.0 & 63.0 & 59.2 & 61.7 & 72.3  & 58.8 \\
LLaVA-NeXT~\cite{li2024llavanext-strong} & & 72B & 52.0 & 51.5 & 49.4 & 51.5 & 50.8 & 55.3 & 56.8 & 59.3  & 53.3 \\
\midrule

\multicolumn{12}{c}{\textbf{Open-Source Reasoning MLLMs}} \\
\midrule
LLaVA-CoT-7B~\cite{llava_cot} & & 7B &
51.5 & 50.8 & 51.5 & 51.3 & 54.8 & 55.3 & 53.9 & 54.8  & 53.0 \\

Mulberry-7B~\cite{yao2024mulberry} & & 7B &50.4 & 50.2 & 50.5 & 51.2 & 63.4 & 54.5 & 53.4 & 56.0 & 53.7 \\
\midrule
\multicolumn{12}{c}{\textbf{Our Model}} \\
\midrule
\textbf{ThinkFake (Ours)} & & 7B & \textbf{92.5} & \textbf{93.1} & \textbf{95.3} & \textbf{73.1} & \textbf{87.4} & \textbf{93.6} & 66.2 & 70.8 & \textbf{84.0}  \\
\bottomrule
\end{tabular}
}
\caption{Performance comparison in Acc. $(\%)$ on selected generative models. For ThinkFake, except for SD v1.4 (used in training), all others are unseen test sets. The best results are highlighted in bold.}
\label{tab:custom-metrics}
\vspace{-10pt}

\end{table*}

\noindent\textbf{Agentic Reward ($f_{agentic}$).} The agentic reward corresponds to the ``Auxiliary Tool Inspection" step in our pipeline. We incorporated state-of-the-art models as expert agents to enhance the inspection process from multiple perspectives: UnivFD~\cite{ojha2023univfd} for high-level semantic artifact analysis, NPR~\cite{npr} for low-level frequency-based artifact features, and AIDE~\cite{yan2025aide} as a dual-stream agent addressing both levels. The outputs of these agents serve as ground-truth labels for their respective inspection tasks. Rewards are computed via cross-entropy between these labels and the model's predictions. Leveraging supervision from expert agents enables the model to achieve more robust and comprehensive artifact detection. \textit{Further details are provided in the Appendix.~\ref{app:reward_function}.}

\subsection{Training Process}\label{sec:training}
\noindent\textbf{Data Preparation.} We follow the GenImage protocol~\cite{zhu2024genimage} for dataset construction. Real images are sourced from ImageNet, while fake images are generated using SD v1.4 based on 1k ImageNet labels. To ensure balance, we uniformly sample each class to create a sub-dataset with equal numbers of real and fake images. This sub-dataset is split into two parts: one for cold-start supervised fine-tuning (SFT-set) and one for reinforcement learning (RL-set). \textbf{1) SFT-set:} Since the original data contains only binary labels, we design a data annotation pipeline using a commercial MLLM (e.g., Gemini-1.4-pro) to generate reasoning paths and attribute-level answers. The ground-truth label is included in the prompt to reduce hallucinations. We further apply rule-based filtering to remove low-quality responses, resulting in 638 high-quality samples for SFT training. \textbf{2) RL-set:} RL training uses only binary labels, so no additional annotation is needed. We use 5,000 samples (2,500 real and 2,500 fake) as the RL training set.

\noindent\textbf{GRPO.} We fine-tune the pre-trained Qwen2.5-VL-7B-Instruct model using the proposed reward signals to guide policy learning. For optimization, we adopt the Group Relative Policy Optimization (GRPO) algorithm~\cite{GRPO, deepseekr1}, which eliminates the need for a separate critic network by leveraging group-level reward statistics to estimate baselines, thereby reducing computational overhead.

Given a question $q$, GRPO draws a set of $G$ responses ${o_1, o_2, \ldots, o_G}$ from the old policy $\pi_{\theta_{\text{old}}}$ and updates the current policy $\pi_\theta$ by maximizing the following objective:

\begin{equation}
\begin{aligned}
\mathcal{J}_{\text{GRPO}}(\theta) = \mathbb{E}_{q, \{o_i\}} \Bigg[
& \frac{1}{G} \sum_{i=1}^{G} \Big( 
\min \Big( 
\frac{\pi_\theta(o_i|q)}{\pi_{\theta_{\text{old}}}(o_i|q)} A_i, \\
& \quad\text{clip} \Big( \frac{\pi_\theta(o_i|q)}{\pi_{\theta_{\text{old}}}(o_i|q)}, 1 - \epsilon, 1 + \epsilon \Big) A_i 
\Big) \\
& - \beta \, \mathbb{D}_{\text{KL}}(\pi_\theta \,\|\, \pi_{\text{ref}}) \Big)
\Bigg],
\end{aligned}
\label{eq:grpo_objective}
\end{equation}

\noindent where the hyperparameters $\epsilon$ and $\beta$ regulate the clipping threshold and the KL regularization strength, respectively. The KL-divergence term is computed as:
\begin{equation}
\mathbb{D}_{\text{KL}}(\pi_\theta \| \pi_{\text{ref}}) = 
\frac{\pi_{\text{ref}}(o_i|q)}{\pi_\theta(o_i|q)} - \log \frac{\pi_{\text{ref}}(o_i|q)}{\pi_\theta(o_i|q)} - 1,
\label{eq:kl_divergence}
\end{equation}
\noindent and the normalized advantage $A_i$ is derived from the group-wise reward set ${r_1, r_2, \ldots, r_G}$ as:
\begin{equation} A_i = \frac{r_i - \text{mean}(\{r_1, r_2, \ldots, r_G\}))}{\text{std}(\{r_1, r_2, \ldots, r_G\}))}. \label{eq:advantage} \end{equation}\\
\noindent\textbf{Training ThinkFake.}
Overall, our ThinkFake approach begins with supervised fine-tuning (SFT) using the SFT-set for cold-start initialization. The high-quality SFT-set data enables the reasoning MLLM to develop initial capabilities for AI-generated image detection while aligning with the desired output format. Subsequently, we employ reinforcement learning with the RL-set using GRPO to enhance the MLLM’s generalization ability. At this stage, we utilize four carefully designed reward functions, as described in Sec.~\ref{sec:reward}, to guide the model. Two format reward functions constrain the model’s reasoning and output structure. The ``Accuracy Reward'' improves yes/no discrimination, while the ``Agentic Reward'' enables multi-perspective evaluation for better performance and interpretability. \textit{Full data pipeline and training details are in Appendix~\ref{app:training_process}.}

\begin{table*}[ht]
\centering
\begin{tabular}{lcccccccc}
\toprule
\textbf{} & Overall & Scene & Animal & Person & Object & Medicine & Doc & Satellite \\
\midrule
Human & 27.3 & 24.0 & 25.8 & 19.9 & 26.9 & 26.1 & 22.1 & 39.3 \\
AIDE~\cite{yan2025aide} & 63.1 & - & \textbf{89.9} & 62.5 & \textbf{96.5} & 53.4 & 49.7 & 39.3 \\
\midrule
VILA1.5-40B & 48.8 & 53.7 & 39.3 & 50.0 & 33.4 & 52.5 & 59.9 & 50.6 \\
InternVL2-40B & 49.6 & 55.7 & 37.3 & 59.2 & 34.8 & 55.5 & 64.8 & 40.8 \\
Qwen2-VL-72B & 53.2 & 55.9 & 43.4 & 66.9 & 38.0 & 55.9 & 73.7 & 38.2 \\
LLaVA-OneVision-72b & 46.3 & 54.7 & 31.6 & 53.1 & 27.8 & 52.1 & \textbf{67.9} & 36.6 \\
\midrule
Claude-3.5-Sonnet & 53.6 & 51.6 & 51.6 & 55.2 & 51.4 & 51.9 & 59.1 & 50.9 \\
Gemini-1.5-Pro & 43.5 & 53.7 & 35.7 & 51.5 & 30.3 & 50.0 & 47.2 & 38.1 \\
GPT-4o & 63.4 & 70.1 & 69.7 & \textbf{84.4} & 70.3 & 54.3 & 60.1 & 45.0 \\
\midrule
\textbf{ThinkFake (Ours)} & \textbf{75.4} & \textbf{78.5} & 87.2 & 71.7 & 89.4 & \textbf{64.5} & 54.2 & \textbf{82.2} \\
\bottomrule
\end{tabular}

\caption{Quantitative results in Acc. $(\%)$ on the LOKI~\cite{ye2024loki} benchmark. All categories are unseen zero-shot test cases. The best results are highlighted in bold.}
\label{tab:loki_bench}

\end{table*}

\section{Experiments}
\label{sec:exp}
\subsection{Experimental Settings}
\noindent\textbf{Implementation Details.}
\label{sec:exp_details}
We initialize the reasoning MLLM $\mathcal{F}_{\theta}$ using Qwen2.5-VL-7B-Instruct~\cite{bai2025qwen2}. The cold-start phase comprises 58 iterations with a batch size of 40 and a learning rate of $5\mathrm{e}{-6}$, implemented via LLaMA-Factory~\footnote{\url{https://github.com/hiyouga/LLaMA-Factory}}. This is followed by GRPO training for 1,250 iterations with a batch size of 256, a learning rate of $1\mathrm{e}{-6}$, and 8 generations, conducted using the VLM-R1~\footnote{\url{https://github.com/om-ai-lab/VLM-R1}} framework.
To implement the expert agents—UnivFD~\cite{ojha2023univfd}, NPR~\cite{npr}, and AIDE~\cite{yan2025aide}—we adopt their official open-source code and pre-trained weights.
All experiments are conducted on 8 NVIDIA H200 GPUs.

\noindent\textbf{Datasets.}
\label{sec:exp_dataset}
As noted in Sec.~\ref{sec:training}, we adhere to the GenImage protocol~\cite{zhu2024genimage} but employ our constructed SFT-set and RL-set sub-training sets for cold-start SFT and RL training, respectively. Testing follows the GenImage benchmark, with training performed on SD v1.4 and generalization evaluated on unseen models, including SD v1.5, Midjourney, ADM, GLIDE, Wukong, VQDM, and BigGAN.

\noindent\textbf{Evaluation Metrics.} Following prior work~\cite{wang2019cnnspot, yan2025aide}, we report classification accuracy (Acc.) averaged over both real and AI-generated images. 


\subsection{Can MLLMs Handle AI-generated Image Detection?}
We first investigate whether existing state-of-the-art multimodal large language models (MLLMs) possess AI-generated image detection capabilities. We evaluated open-source general MLLMs, including the Qwen2.5-VL series, InternVL-2.5, and LLaVA-Next, as well as open-source reasoning MLLMs such as LLaVA-CoT and Mulberry. Additionally, we assessed closed-source commercial models like ChatGPT-4o, Gemini-2.0-Flash, and the widely recognized reasoning model ChatGPT-o3. The results, presented in Tab.~\ref{tab:custom-metrics}, reveal that these large-scale MLLMs, including reasoning models, fail to distinguish real images from AI-generated ones, underscoring the importance and urgency of this task. In contrast, our ThinkFake model demonstrates superior detection capabilities, performing competitively despite having significantly fewer parameters. These results highlight ThinkFake’s strong reasoning capabilities, enabling it to generalize to unseen categories and effectively handle complex detection and explanation tasks.

\begin{table}[t]
\centering

\begin{tabular}{lc|c}
\toprule
\textbf{Method} & \textbf{Reasoning} &\textbf{Mean Acc.} \\
\midrule
CNNSpot & & 62.6 \\
AntifakePrompt & & 76.9 \\
UnivFD & & 72.3 \\
NPR & & 59.6 \\
AIDE & & 81.7 \\
\midrule
\textbf{ThinkFake (Ours)} & \checkmark & \textbf{84.0} \\
\bottomrule
\end{tabular}
\caption{\textbf{Comparison of State-of-the-Art Methods:}
All baselines are re-trained on our training set to ensure a fair comparison, following the GenImage evaluation protocol.}
\label{tab:compare_sotas}
\vspace{-5pt}
\end{table}

\subsection{Comparison with Other SOTA Methods}
To assess detection performance, we compare our proposed method with SOTA method~\cite{wang2019cnnspot, yan2025aide, npr, ojha2023univfd, antifakeprompt}. For a fair comparison, all baseline detectors are re-trained using our experimental configuration, which adheres to the GenImage protocol and trains solely on the sub-training set. Evaluation is conducted on unseen generators, as described in Sec.\ref{sec:exp_dataset}.
In this experiment, we report the mean image-level accuracy (Mean Acc.) across all test sets, as shown in Tab.~\ref{tab:compare_sotas}. Our ThinkFake model achieves strong detection performance. Unlike the other SOTA methods, which are limited to binary classification, our method also generates interpretable reasoning, further demonstrating the effectiveness of our reinforcement learning framework in guiding the model’s decision-making process for AI-generated image detection. \textit{A comprehensive evaluation will be presented in the Appendix.~\ref{app:sota_compare}.}

\subsection{Generalization in Real-World Scenarios}
In Tab.~\ref{tab:loki_bench}, we assess the generalization capability of our ThinkFake model against the LOKI benchmark~\cite{ye2024loki}. LOKI benchmark is a novel, comprehensive evaluation for MLMMs to detect synthetic data across multimodal, with 18,000 questions across 26 subcategories. We selected the ``Image Judgment Task'', where MLLMs identify synthetic or real images in categories like Scene, Animal, Person, Object, Medicine, Doc, and Satellite. Following the LOKI protocol, we zero-shot tested our ThinkFake model. As shown in Tab.~\ref{tab:loki_bench}, ThinkFake exhibited strong generalization, performing well across unseen categories, with slight performance drops in Doc and Medicine but still competitive results. Notably, ThinkFake exhibits strong and comprehensive robustness under unseen scenarios. Even for rare or uncommon categories, the model maintains high performance, suggesting that its reasoning and explanatory capabilities can effectively generalize to real-world challenges. These results underscore the effectiveness of our approach. \textit{Additional baseline methods and visual examples will be introduced in the Appendix.~\ref{app:loki_benchmark}.}

\begin{table}[t]
\centering
\resizebox{0.45 \textwidth}{!}{
\begin{tabular}{lcc|c}
\toprule
\textbf{Method} & \textbf{Cold Start} & \textbf{GRPO}   & \textbf{Mean Acc.} \\
\midrule
Baseline &  & & 54.1\\
ThinkFake-Zero & & \checkmark  & 72.4 \\
ThinkFake-SFT & \checkmark & & 76.7\\
\textbf{ThinkFake-R1 (Ours)} & \checkmark & \checkmark  & \textbf{84.0} \\
\bottomrule
\end{tabular}
}
\caption{\textbf{Comparison of Training Strategies:}
We compare ThinkFake with various training methods under the GenImage evaluation protocol. The baseline refers to our initial reasoning model, Qwen2.5-VL-7B-Instruct.}
\label{tab:training_strategy}
\end{table}

\begin{table}[t]
\centering

\begin{tabular}{cccc|c}
\toprule
\textbf{$f_{format}$} & \textbf{$f_{json}$} & \textbf{$f_{acc}$} & \textbf{$f_{agentic}$}   & \textbf{Mean Acc.} \\
\midrule
 &  & \checkmark & & \XSolidBrush \\
\checkmark &  & \checkmark & & 74.6 \\
\checkmark & \checkmark & \checkmark &   & 82.6 \\
\checkmark & \checkmark & \checkmark  & \checkmark & \textbf{84.0} \\
\bottomrule
\end{tabular}
\caption{\textbf{Comparison of Reward Functions:}
We evaluate the impact of different reward functions on the performance of ThinkFake under the GenImage evaluation protocol. A \XSolidBrush\ symbol indicates training failure.}
\label{tab:reward_function}

\end{table}

\begin{figure*}[t]
    \centering
    \includegraphics[width=0.95\textwidth]{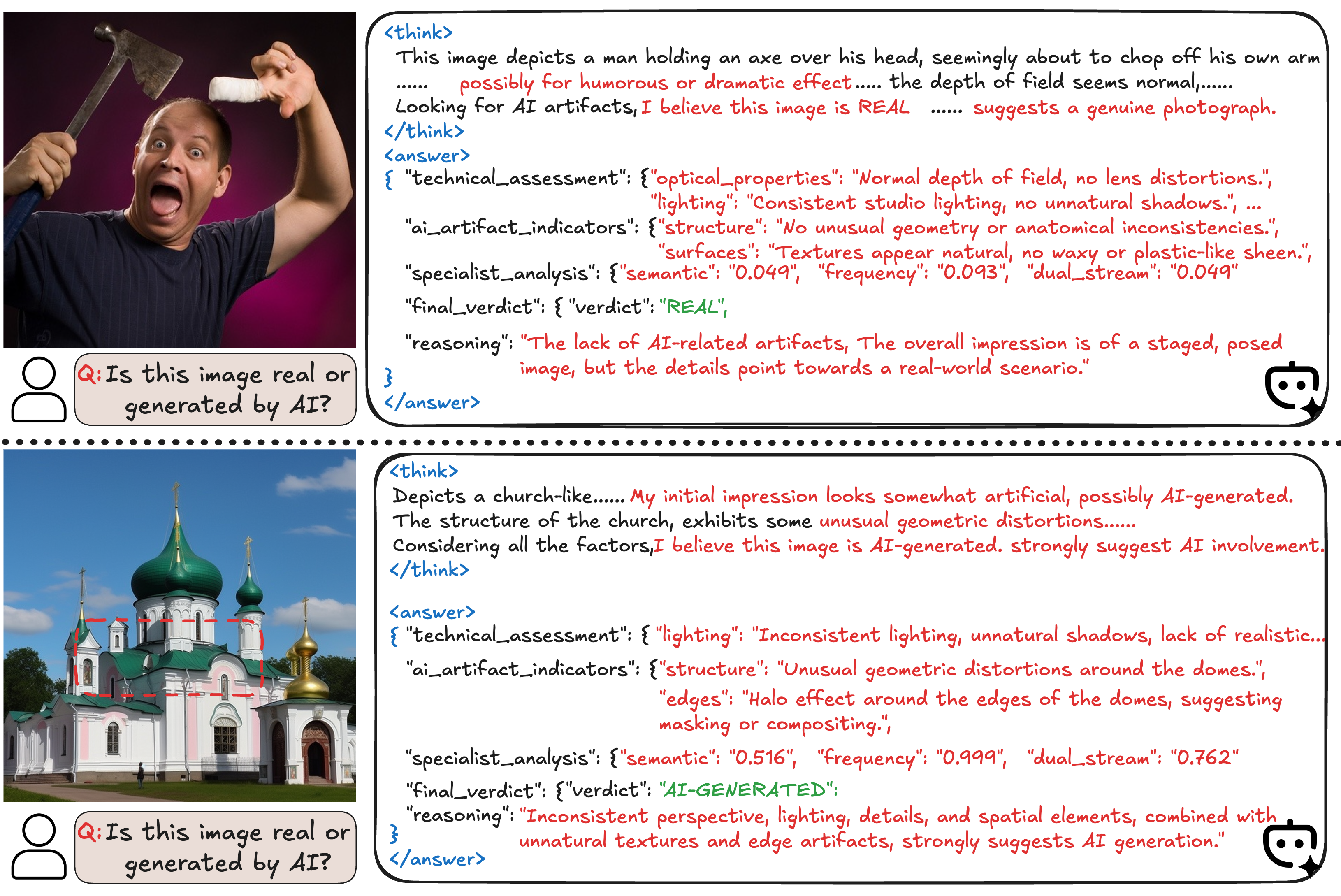}
    \caption{\textbf{Qualitative Results:}
We present qualitative examples demonstrating ThinkFake’s capability in detecting AI-generated images and providing corresponding explanations.}
    \label{fig:qualitative} 
    \vspace{-4pt}
\end{figure*}

\subsection{Ablation Study}
\noindent\textbf{Training Strategy}
Inspired by DeepSeek-R1-Zero~\cite{deepseekr1}, we explored training strategies that guide models toward self-reflected reasoning for AI-generated image detection. Specifically, we followed the DeepSeek-R1-Zero pipeline to train a reasoning MLLM using GRPO with our RL-set, as described in Sec.~\ref{sec:training}. This variant, named ``ThinkFake-Zero'', was trained using only the ``Reasoning Format Reward'' and ``Accuracy Reward'' (Sec.~\ref{sec:reward}) to encourage reasoning behavior. We also implemented the DeepSeek-R1 pipeline, which begins with a cold-start initialization followed by GRPO training. Using the SFT-set, we first obtained ``ThinkFake-SFT'', then applied reinforcement learning to produce the final model, ``ThinkFake-R1''. As shown in Tab.~\ref{tab:compare_sotas}, ThinkFake-Zero improved upon the base model but exhibited limited generalization. This suggests that applying RL alone to enhance reasoning in MLLMs remains challenging. In contrast, ThinkFake-SFT, trained with high-quality supervision, achieved stable output formatting and basic task competency. Subsequent RL fine-tuning led to ThinkFake-R1, which demonstrated strong generalization across diverse scenarios.

\noindent\textbf{Reward Function.}
In Tab.~\ref{tab:reward_function}, we evaluate the impact of different reward functions on the framework’s generalization across all test set. Our results highlight the critical role of carefully designed rewards. The accuracy reward ($f_{acc}$) is key to driving convergence during reinforcement learning, while omitting the reasoning format reward ($f_{format}$) leads to unstructured and unstable outputs. Together, $f_{acc}$ and $f_{format}$ provide a stable foundation for effective training. Adding the JSON structure reward ($f_{json}$) boosts performance by enforcing outputs aligned with the detection pipeline. Incorporating the agentic reward ($f_{agentic}$), guided by expert agents, further enhances reasoning and enables robust, comprehensive detection.

\subsection{Qualitative Results}
Fig.~\ref{fig:qualitative} showcases the detection and explanation results generated by ThinkFake on both real and AI-generated images. Our method provides detailed thinking processes, diverse perspectives of judgment, and intuitive, human-friendly explanations, clearly demonstrating its effectiveness. \textit{Additional qualitative examples are provided in the Appendix.~\ref{app:qualitative}.}

\section{Conclusions}
\label{sec:conclusion}
In this paper, we introduced ThinkFake, a reasoning MLLM framework for detecting AI-generated images. By integrating a forgery reasoning prompt, GRPO reinforcement learning, and carefully designed reward functions, our method achieves both accurate classification and interpretable reasoning. Evaluations under the GenImage protocol show that ThinkFake surpasses existing state-of-the-art detectors. Moreover, the model generalizes effectively to challenging real-world LOKI benchmarks, highlighting its robustness and practical value. These results highlight the potential of reasoning-based methods for trustworthy AI detection. ThinkFake offers a strong foundation for future research in explainable and generalizable media forensics.

\bibliography{aaai2026}

\clearpage

\clearpage
\appendix
\setcounter{secnumdepth}{1}
\renewcommand{\algorithmicrequire}
{\textbf{Input:}}  
\renewcommand{\algorithmicensure}{\textbf{Output:}} 

\section{Reward Functions}
\label{app:reward_function}
We carefully design four reward functions as follows:

\noindent\textbf{Reasoning Format Reward ($f_{format}$).} Pseudocode is provided in Algorithm~\ref{alg:reasoning_format}.

\noindent\textbf{JSON Format Reward ($f_{json}$).} Pseudocode is provided in Algorithm~\ref{alg:json_format}.

\noindent\textbf{Accuracy Reward ($f_{acc}$).} Pseudocode is provided in Algorithm~\ref{alg:acc_reward}.

\noindent\textbf{Agentic Reward ($f_{agentic}$).} Pseudocode is provided in Algorithm~\ref{alg:agentic_reward}.

\section{Training Process Details}
\label{app:training_process}
As described in Section~\ref{sec:training}, we follow the GenImage protocol for data preparation. After obtaining the full GenImage~\cite{zhu2024genimage} training set, we uniformly sample 6,000 examples, consisting of 3,000 real and 3,000 fake images. This sampled data is then split into two subsets: 1,000 samples for supervised fine-tuning (SFT-set) and 5,000 samples for reinforcement learning (RL-set).

\begin{figure*}[t]
    \centering
    \includegraphics[width=\textwidth]{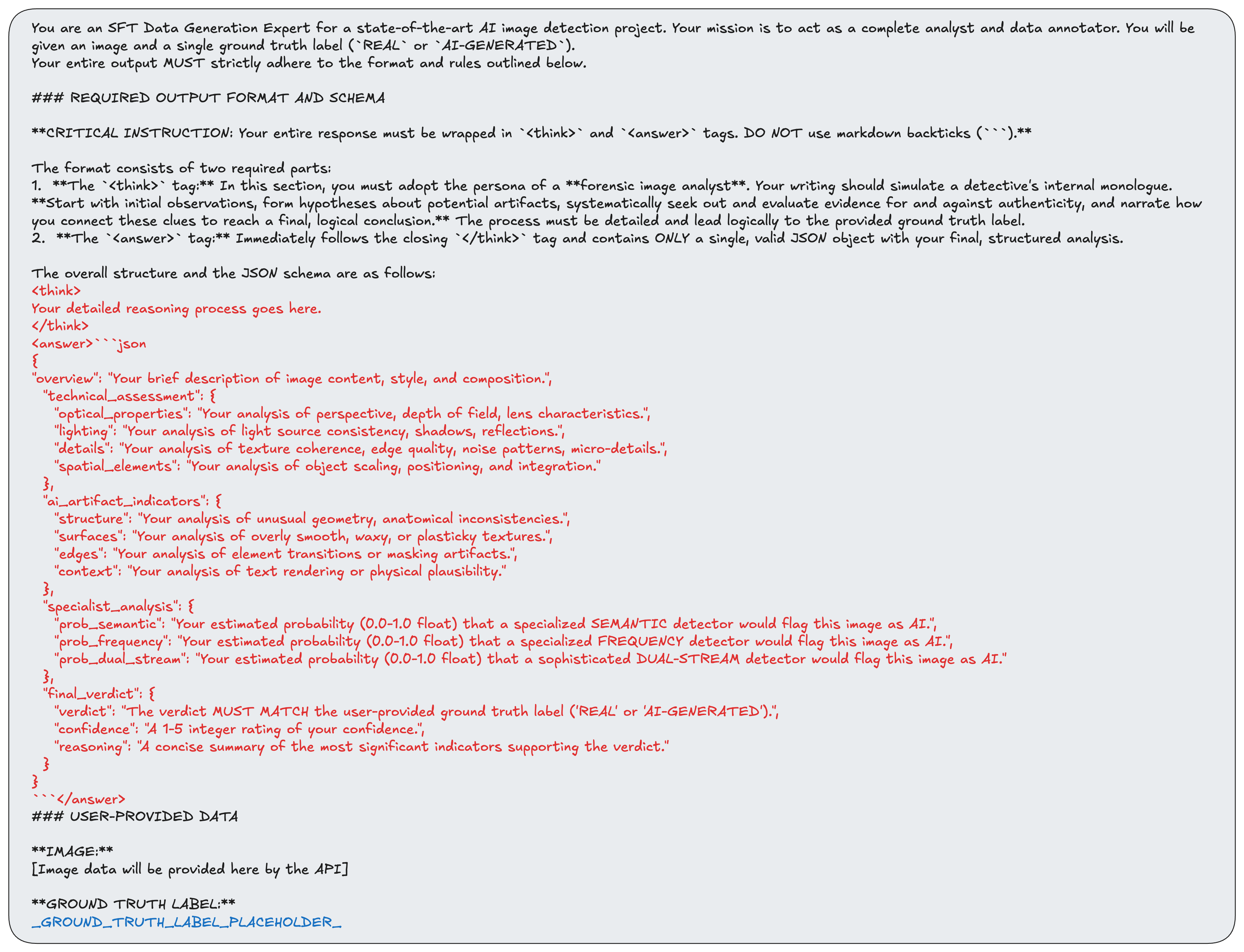}
    \caption{\textbf{SFT-set Preparation Prompt:} We adapt the  ``Forgery Reasoning Prompt'' as the base and design a ground-truth injection mechanism to guide the MLLM in generating high-quality reasoning without hallucination.}
    \label{fig:sft_pipeline} 
\end{figure*}

The original dataset provides only binary labels (real/fake). To construct a high-quality SFT-set, we design a three-stage data preparation pipeline: \textbf{1)} We adapt the ``Forgery Reasoning Prompt'' as described in Sec.~\ref{sec:pipeline} and inject prior ground-truth information into the prompt to reduce hallucinations (see Fig.~\ref{fig:sft_pipeline} for full prompt example). \textbf{2)} Using a commercial MLLM (e.g., Gemini-1.4-pro), we generate attribute-level explanations based on the prompts and corresponding images. \textbf{3)} A rule-based post-processing step filters out low-quality outputs to ensure consistency and accuracy. Through this process, we obtain 638 high-quality annotated samples, which form the final SFT-set. This carefully designed pipeline serves as a critical foundation for the strong performance of our model.

As described in Section~\ref{sec:exp_details}, the training of our ThinkFake model consists of two stages. The first stage is a cold-start initialization, where we perform supervised fine-tuning using the SFT-set under the LLaMA-Factory~\footnote{\url{https://github.com/hiyouga/LLaMA-Factory}} framework to align the model's output with the AI-generated image detection task. In the second stage, we adopt the VLM-R1~\footnote{\url{https://github.com/om-ai-lab/VLM-R1}} framework and train the model on the RL-set using the GRPO reinforcement learning algorithm. During this phase, we employ four reward functions—``Reasoning Format Reward'', ``JSON Format Reward'', ``Accuracy Reward'', and ``Agentic Reward''—to guide and optimize the training process.

\begin{algorithm}[t]
\caption{Reasoning Format Reward}
\label{alg:reasoning_format}
\begin{algorithmic}[1]
    \STATE \textbf{Input:} $Completions$: A list of content generated by the model.
    \STATE \textbf{Output:} $Rewards$: A list of reward values.
    \STATE \textbf{Initialize} $Rewards \leftarrow []$
    \STATE \textbf{Define} $Pattern \leftarrow$ \lt think\gt...\lt /think\gt~\lt answer\gt...\lt/answer\gt
    \FORALL{$C$ in $Completions$}
        \IF{$C$ matches the full $Pattern$}
            \STATE Append $1.0$ to $Rewards$
        \ELSE
            \STATE Append $0.0$ to $Rewards$
        \ENDIF
    \ENDFOR
    \STATE \textbf{return} $Rewards$
\end{algorithmic}
\end{algorithm}

\begin{algorithm}[t]
\caption{JSON Format Reward}
\label{alg:json_format}
\begin{algorithmic}[1]
    \STATE \textbf{Input:} $Completions$: A list of content generated by the model.
    \STATE \textbf{Output:} $Rewards$: A list of reward values.
    \STATE \textbf{Initialize} $Rewards \leftarrow []$
    \FORALL{$C$ in $Completions$} 
        \STATE $R \leftarrow 0.0$ 
        \STATE \textbf{try} 
        \STATE \quad  $ExtractedJSON \leftarrow$ Extract from $C$ within \texttt{```json...```} 
            \begin{ALC@g}
                \IF {$ExtractedJSON$ is not empty} 
                    \STATE Parse $ExtractedJSON$ 
                    \STATE $R \leftarrow 1.0$ 
                \ENDIF 
            \end{ALC@g}
            \STATE \textbf{catch} JSONParseException 
            \STATE \quad $R \leftarrow 0.0$ 
        \STATE \textbf{end try} 
        \STATE Append $R$ to $Rewards$ 
    \ENDFOR 
    \STATE \textbf{return} $Rewards$
\end{algorithmic}
\end{algorithm}

\begin{algorithm}[t]
\caption{Accuracy Reward}
\label{alg:acc_reward}
\begin{algorithmic}[1]
    \STATE \textbf{Input:} $Completions$: A list of content generated by the model. $Solutions$: The ground truth label for each completion.
    \STATE \textbf{Output:} $Rewards$: A list of calculated reward values.
    \STATE \textbf{Initialize} $Rewards \leftarrow []$
    \FORALL{($C, S$) in zip($Completions$, $Solutions$)}
        \STATE $R \leftarrow -1.0$
        \STATE \textbf{try}
        \STATE \quad $ExtractedJSON \leftarrow$ Extract JSON from $C$
        \STATE \quad $ParsedAnswer \leftarrow$ Parse $ExtractedJSON$
        \STATE \quad $Verdict \leftarrow ParsedAnswer.final\_verdict.verdict$
        \begin{ALC@g}
            \IF{
                ($S = 0$ \textbf{and} $Verdict = \text{``REAL''}$) \textbf{or} \newline
                \quad ($S = 1$ \textbf{and} $Verdict = \text{``AI-GENERATED''}$)
            }
                \STATE \quad $R \leftarrow 1.0$
            \ELSE
                \STATE \quad $R \leftarrow 0.0$ 
            \ENDIF
        \end{ALC@g}
        \STATE \textbf{catch} JSONParseException
        \STATE \quad $R \leftarrow -1.0$ 
        \STATE \textbf{end try}
        \STATE Append $R$ to $Rewards$
    \ENDFOR
    \STATE \textbf{return} $Rewards$
\end{algorithmic}
\end{algorithm}

\begin{algorithm}[t]
\caption{Agentic Reward}
\label{alg:agentic_reward}
\begin{algorithmic}[1]
    \STATE \textbf{Input:} $Completions$: A list of content generated by the model. $Solutions$: The ground truth label for each completion. $AgenticSolutions$: The ground truth probability distribution data.
    \STATE \textbf{Output:} $Rewards$: A list of calculated reward values.
    \STATE \textbf{Initialize} $Rewards \leftarrow []$
    \STATE $ProbKeys \leftarrow [prob_{semantic}, prob_{frequency}, prob_{dual}]$.
    \FORALL{($C, S, A$) in zip($Completions$, $Solutions$, $AgenticSolutions$)}
        \STATE $R \leftarrow -1.0$
        \STATE \textbf{try}
        \STATE \quad $ExtractedJSON \leftarrow$ Extract JSON from $C$
        \STATE \quad $ParsedAnswer \leftarrow$ Parse $ExtractedJSON$
        \STATE \quad $Verdict \leftarrow ParsedAnswer.final\_verdict.verdict$
        \begin{ALC@g}
            \IF{
                ($S = 0$ \textbf{and} $Verdict = \text{``REAL''}$) \textbf{or} \newline
                \quad ($S = 1$ \textbf{and} $Verdict = \text{`AI-GENERATED''}$)
            }
                \STATE \quad \textbf{Initialize} $Losses \leftarrow []$
                \STATE \quad $PredProbs \leftarrow ParsedAnswer.specialist\_analysis$
                \FORALL{$key$ in $ProbKeys$}
                    \STATE \quad $p_{\text{pred}} \leftarrow PredProbs[key]$
                    \STATE \quad $p_{\text{gt}} \leftarrow A[key]$
                    \STATE \quad $\mathcal{L}_{\text{BCE}} \leftarrow \text{BinaryCrossEntropy}(p_{\text{pred}}, p_{\text{gt}})$
                    \STATE \quad Append $\mathcal{L}_{\text{BCE}}$ to $Losses$
                \ENDFOR
                \STATE \quad $\mathcal{L}_{\text{avg}} \leftarrow \text{Mean}(Losses)$
                \STATE \quad $R \leftarrow e^{-\mathcal{L}_{\text{avg}}}$
            \ELSE
                \STATE \quad $R \leftarrow 0.0$
            \ENDIF
        \end{ALC@g}
        \STATE \textbf{catch} JSONParseException
        \STATE \quad $R \leftarrow -1.0$ 
        \STATE \textbf{end try}
        \STATE Append $R$ to $Rewards$
    \ENDFOR
    \STATE \textbf{return} $Rewards$
\end{algorithmic}
\end{algorithm}

\section{Compare with SOTA Methods}
\label{app:sota_compare}
We compare our method with several state-of-the-art approaches, including CNNSpot~\cite{wang2019cnnspot}, AntifakePrompt~\cite{antifakeprompt}, UnivFD~\cite{ojha2023univfd}, NPR~\cite{npr}, and AIDE~\cite{yan2025aide}. For fair comparison, we reproduce their results using their official open-source codes and retrain all models on the same sub-training set constructed under the GenImage~\cite{zhu2024genimage} protocol, as described in Section~\ref{sec:training}. Evaluation is conducted on a cross test-set to assess generalization. Note that only SD v1.4 overlaps with the training domain, while all other test sets contain unseen categories.

The full cross-evaluation results are reported in Table~\ref{tab:full_sota}, with accuracy (ACC. \%) as the evaluation metric. Despite being trained on a limited dataset, our ThinkFake model demonstrates strong and consistent generalization across diverse test domains. This suggests that the model effectively performs reasoning-based detection. Furthermore, unlike other detectors that provide only binary predictions, ThinkFake generates interpretable, structured outputs—highlighting its effectiveness in explainable AI-generated image detection.

\section{LOKI Benchmark}
\label{app:loki_benchmark}
Generalization has long been one of the key challenges in AI-generated image detection, limiting the practical applicability of previous methods. The Image modality of the LOKI~\cite{ye2024loki} benchmark offers a valuable setting for evaluating generalization in real-world scenarios. This evaluation dataset contains over 2,200 images from seven real-world subcategories (e.g., Scene, Animal, Person, Object, Medicine, Document, Satellite), collected through existing datasets, internet sources, and newly synthesized data. Image generation methods include FLUX, Midjourney, Stable Diffusion, and ten additional techniques to ensure high quality and diversity. The task requires the model to determine whether each image is real or synthetic, presenting a challenging evaluation of generalization to complex and diverse scenarios.

\begin{figure*}[t]
    \centering
    \includegraphics[width=\textwidth]{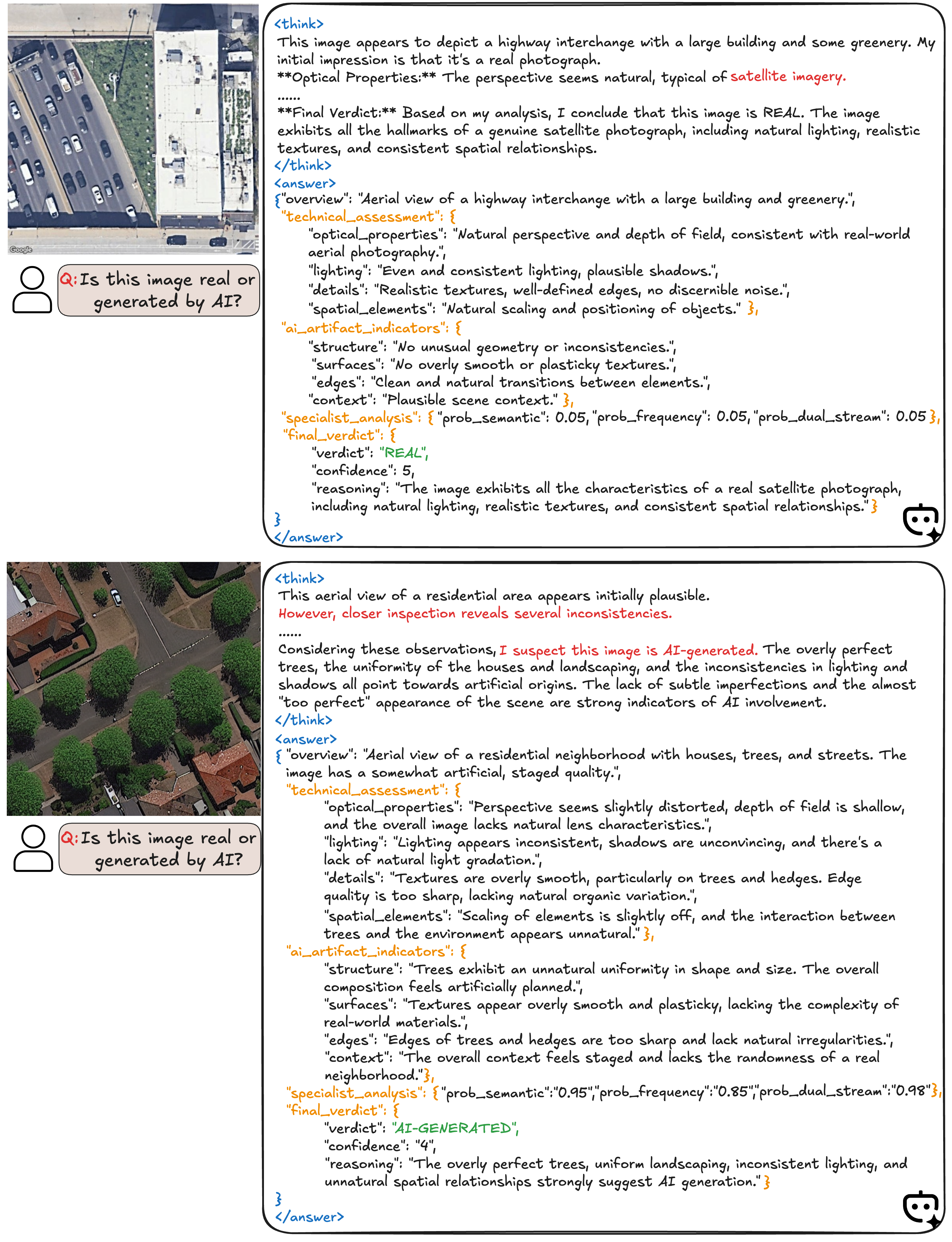}
    \caption{\textbf{Qualitative Results:} We present qualitative examples of ThinkFake on the ``Satellite'' category from the LOKI benchmark. Despite the rarity of such images, ThinkFake consistently provides high-quality and robust predictions along with clear explanations.}
    \label{fig:satellite} 
\end{figure*}

Following the LOKI protocol and its comprehensive evaluation results, we conduct a zero-shot evaluation of our ThinkFake method. The results are presented in Table~\ref{app:loki_benchmark}. Our method demonstrates unprecedented detection performance and robustness across diverse categories. While performance slightly decreases in rare and domain-specific categories such as ``Medicine'' and ``Document'', the overall results highlight the effectiveness of our framework. We attribute this success to the combination of our reasoning-based design and the generalization capability enabled by reinforcement learning, which allows the model to reason effectively even in unseen scenarios. Notably, although ``Satellite'' is also a rare category, our ThinkFake model achieves outstanding performance. As shown in Fig.~\ref{fig:satellite}, the model is able to think accurately and provide interpretable answers. This success can be attributed not only to the strong knowledge base of the underlying base MLLM, but more importantly, to our carefully designed reasoning pipeline. This pipeline guides the model to analyze, differentiate, and summarize before making a final decision with explanation—demonstrating practical applicability in real-world scenarios.

\section{Qualitative Results}
\label{app:qualitative}
\begin{figure*}[t]
    \centering
    \includegraphics[width=\textwidth]{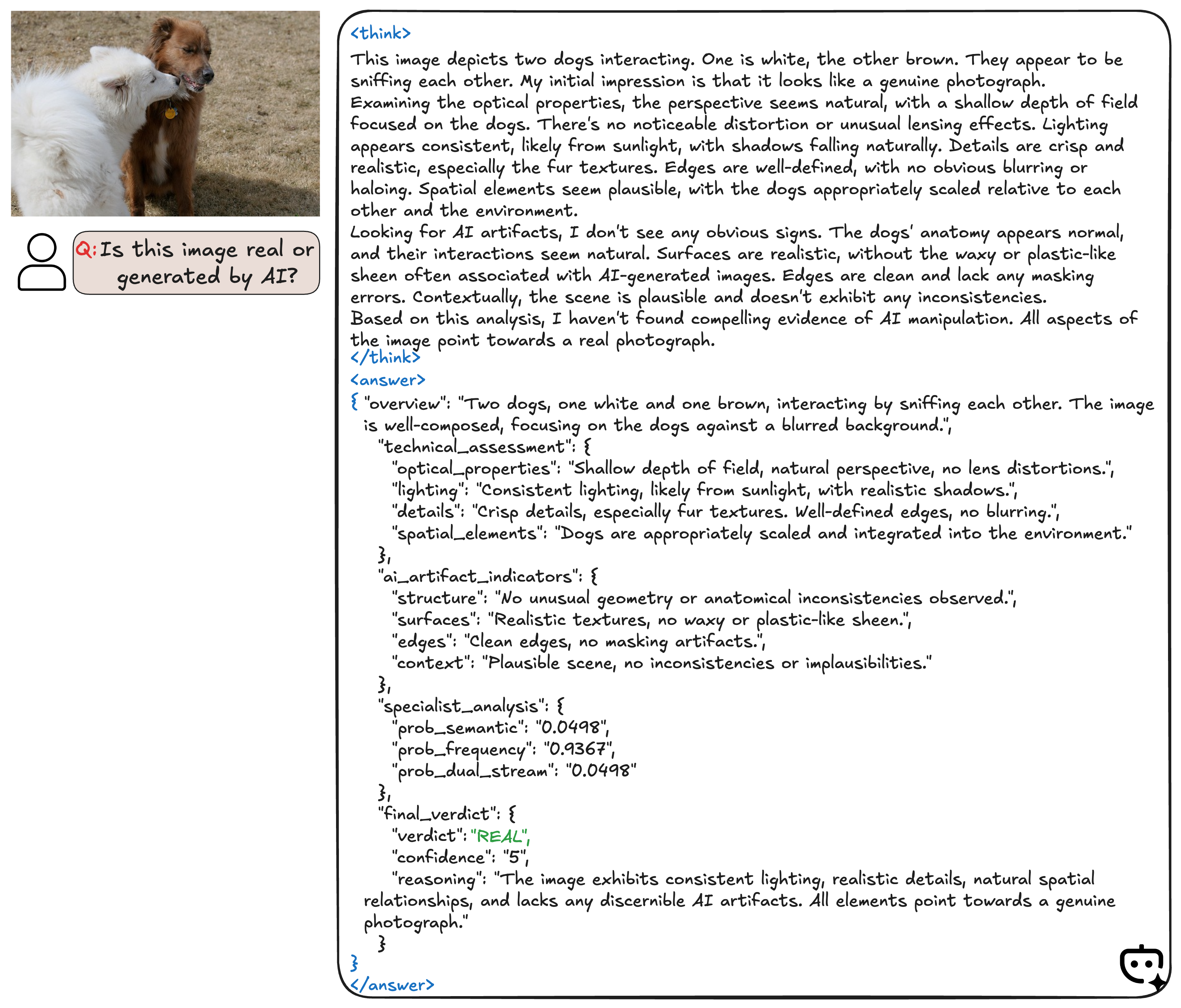}
    \caption{\textbf{Qualitative Result:} Detection and explanation of real image by ThinkFake.}
    \label{fig:qual_real} 
\end{figure*}

\begin{figure*}[t]
    \centering
    \includegraphics[width=\textwidth]{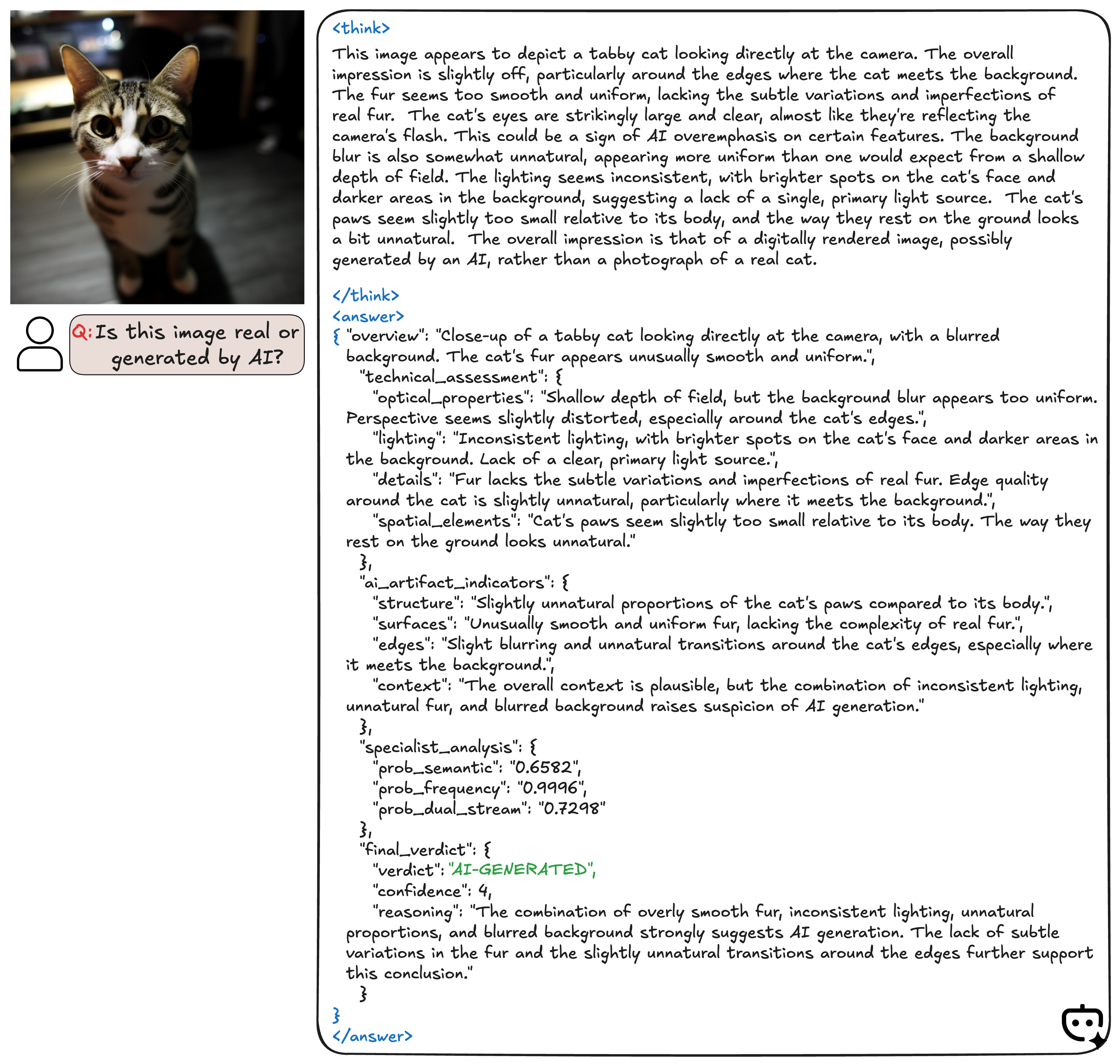}
    \caption{\textbf{Qualitative Result:} Detection and explanation of AI-generated image by ThinkFake.}
    \label{fig:qual_fake} 
\end{figure*}

We provide additional qualitative examples in Fig.\ref{fig:qual_real} and Fig.\ref{fig:qual_fake}, showcasing the complete outputs generated by ThinkFake. These outputs include the full step-by-step reasoning process, along with clearly structured answers. ThinkFake demonstrates not only accurate detection but also user-friendly and interpretable explanations.

\section{Limitation and Future Work}
\label{app:limit_future}
\noindent \textbf{Limitation.}
Our proposed ThinkFake framework establishes a complete detection pipeline and leverages reinforcement learning with carefully designed reward functions for training. Unlike previous methods~\cite{fakereasoning, zhou2025aigi}, which rely heavily on large-scale supervised fine-tuning (SFT) data and techniques such as prompt optimization or in-context learning to simulate reasoning behavior, ThinkFake achieves more generalized and effective results using only a small amount of SFT data and coarse-labeled RL data. Through a well-designed pipeline, ThinkFake enables the model to perform self-reasoning and structured detection. 

However, since reinforcement learning is applied directly to the base MLLM, we freeze the visual encoder and fully fine-tune the LLM component to reduce computational cost. Despite this, a key limitation of ThinkFake is its requirement for substantial computational resources during training. Additionally, because the model relies on reasoning to produce decisions, real-time detection remains challenging during inference.

\noindent \textbf{Future Work.}
In this paper, ThinkFake has demonstrated strong detection performance, interpretability, and robust generalization to real-world challenges. However, for end users, visual cues are often more intuitive than textual explanations. Future work could integrate segmentation models to map ThinkFake’s reasoning outcomes directly onto the image, highlighting relevant regions. Combining visual annotations with textual reasoning may lead to more human-friendly and understandable explanations.

As AI-generated media continues to grow, so do the associated risks. We believe that detection technologies must evolve alongside generation techniques to ensure a safer and more trustworthy digital future.

\begin{table*}[t]
\centering
\begin{tabular}{lccccccccc}
\toprule
\textbf{Method}  &
{Midjourney} & {SD v1.4} & {SD v1.5} &
{ADM} & {GLIDE} & {Wukong} &
{VQDM} & {BigGAN}  & {\textit{Mean}} \\
\midrule
CNNSpot & 49.9 & 84.9 & 85.1 & 50.6 & 53.2 & 70.0 & 51.9 & 55.2 & 62.6  \\
AntifakePrompt & \underline{77.1} & 81.6 & 81.4 & \underline{72.3} & \underline{82.1} &  80.4 & \underline{69.8} & 70.4 & 76.9  \\
UnivFD & 76.6 & 80.5 & 80.6 & 58.5 & 78.9 & 73.9 & 53.1 & \textbf{76.7} & 72.3  \\
NPR & 61.1 & 61.3 & 59.2 & 55.9 & 61.4 & 60.0 & 58.2 & 60.1 & 59.6  \\
AIDE & 73.6 & \textbf{98.7} & \textbf{99.3} & 64.7 & 81.6 & \textbf{95.6} & \textbf{73.6} & 67.0 & \underline{81.7}  \\
\midrule
\textbf{ThinkFake (Ours)} & \textbf{92.5} & \underline{93.1} & \underline{95.3} & \textbf{73.1} & \textbf{87.4} & \underline{93.6} & 66.2 & \underline{70.8} & \textbf{84.0}  \\
\bottomrule
\end{tabular}
\caption{\textbf{Comparison of State-of-the-Art Methods:}
All baselines are re-trained on our training set to ensure a fair comparison, following the GenImage evaluation protocol. Performance comparison in Acc. $(\%)$}
\label{tab:full_sota}
\end{table*}

\begin{table*}[ht]
\centering
\resizebox{\textwidth}{!}{
\begin{tabular}{lcccccccc}
\toprule
\textbf{} & \textbf{Overall} & \textbf{Scene} & \textbf{Animal} & \textbf{Person} & \textbf{Object} & \textbf{Medicine} & \textbf{Doc} & \textbf{Satellite} \\
\midrule
Random Choice & 18.0 & 21.6 & 18.3 & 18.6 & 26.0 & 22.2 & 22.1 & 22.1 \\
Human & 27.3 & 24.0 & 25.8 & 19.9 & 26.9 & 26.1 & 22.1 & 39.3 \\
AIDE~\cite{yan2025aide} & 63.1 & - & \textbf{89.9} & 62.5 & \textbf{96.5} & 53.4 & 49.7 & 39.3 \\
\midrule
MiniCPM-V-2.6 & 44.8 & 52.0 & 34.4 & 53.1 & 31.5 & 53.8 & 51.5 & 38.3 \\
Phi-3.5-Vision & 52.5 & 50.8 & 41.7 & 71.5 & 34.1 & \underline{57.3} & 54.3 & 60.5 \\
LLaVA-OneVision-7B & 49.8 & 59.2 & 41.9 & 58.1 & 37.3 & 52.3 & 53.0 & 50.1 \\
InternLM-XComposer2.5 & 46.4 & 52.7 & 40.0 & 56.7 & 32.5 & 56.1 & 49.8 & 38.2 \\
mPLUG-Owl3-7B & 45.9 & 52.1 & 37.3 & 52.9 & 31.4 & 55.3 & 53.8 & 38.1 \\
Qwen2-VL-7B & 47.8 & 54.7 & 38.9 & 57.9 & 30.3 & 56.0 & 59.6 & 36.9 \\
LongVA-7B & 46.2 & 57.6 & 37.4 & 52.5 & 34.1 & 54.4 & 49.8 & 39.7 \\
Mantis-8B & 54.6 & 54.9 & 52.2 & 54.8 & 53.5 & 53.1 & 51.9 & \underline{63.3} \\
Idefics2-8B & 45.0 & 51.8 & 35.3 & 52.3 & 29.2 & 52.3 & 53.9 & 40.6 \\
InternVL2-8B & 49.7 & 58.8 & 39.4 & 54.4 & 37.8 & 53.9 & 60.2 & 44.2 \\
Llama3-LongVILA-8B & 49.8 & 49.8 & 50.5 & 50.6 & 47.2 & 50.0 & 49.9 & 50.0 \\
VILA1.5-13B & 49.3 & 52.0 & 38.6 & 54.2 & 31.0 & 50.1 & 56.6 & 62.4 \\
InternVL2-26B & 44.3 & 51.6 & 35.4 & 50.8 & 28.2 & 51.3 & 54.4 & 37.6 \\
VILA1.5-40B & 48.8 & 53.7 & 39.3 & 50.0 & 33.4 & 52.5 & 59.9 & 50.6 \\
InternVL2-40B & 49.6 & 55.7 & 37.3 & 59.2 & 34.8 & 55.5 & 64.8 & 40.8 \\
Qwen2-VL-72B & 53.2 & 55.9 & 43.4 & 66.9 & 38.0 & 55.9 & \textbf{73.7} & 38.2 \\
LLaVA-OneVision-72b & 46.3 & 54.7 & 31.6 & 53.1 & 27.8 & 52.1 & \underline{67.9} & 36.6 \\
\midrule
Claude-3.5-Sonnet & 53.6 & 51.6 & 51.6 & 55.2 & 51.4 & 51.9 & 59.1 & 50.9 \\
Gemini-1.5-Pro & 43.5 & 53.7 & 35.7 & 51.5 & 30.3 & 50.0 & 47.2 & 38.1 \\
GPT-4o & \underline{63.4} & \underline{70.1} & 69.7 & \textbf{84.4} & 70.3 & 54.3 & 60.1 & 45.0 \\
\midrule
\textbf{ThinkFake (Ours)} & \textbf{75.4} & \textbf{78.5} & \underline{87.2} & \underline{71.7} & \underline{89.4} & \textbf{64.5} & 54.2 & \textbf{82.2} \\
\bottomrule
\end{tabular}
}
\caption{Accuracy (\%) of different models on judgment questions from the Image modality of the LOKI benchmark. All categories are unseen zero-shot test cases. Best results are in bold, and second-best are underlined.}
\label{tab:full_loki_bench}
\end{table*}

\end{document}